\documentclass[conference]{IEEEtran}
\IEEEoverridecommandlockouts
\usepackage{cite}
\usepackage{amsmath,amssymb,amsfonts}
\usepackage{algorithmic}
\usepackage{graphicx}
\usepackage{textcomp}
\usepackage{xcolor}
\usepackage{url}
\usepackage{hyperref}

\usepackage{multirow}
\usepackage{lipsum}
\usepackage[binary-units=true]{siunitx}
\def\BibTeX{{\rm B\kern-.05em{\sc i\kern-.025em b}\kern-.08em
    T\kern-.1667em\lower.7ex\hbox{E}\kern-.125emX}}
\begin{document}

\title{MLonMCU: TinyML Benchmarking with \\ Fast Retargeting
\thanks{This work was supported in part by the German Federal Ministry of Education and Research (BMBF) within the project Scale4Edge under contract no. 16ME0127.}
}

\author{\IEEEauthorblockN{Philipp van Kempen, Rafael Stahl, Daniel Mueller-Gritschneder, Ulf Schlichtmann}
\IEEEauthorblockA{\textit{Chair of Electronic Design Automation},
\textit{Technical University of Munich}\\
Munich, Germany \\
\{philipp.van-kempen, r.stahl, daniel.mueller, ulf.schlichtmann\}@tum.de}

}

\maketitle

\begin{abstract}
While there exist many ways to deploy machine learning models on microcontrollers, it is non-trivial to choose the optimal combination of frameworks and targets for a given application. Thus, automating the end-to-end benchmarking flow is of high relevance nowadays. A tool called MLonMCU is proposed in this paper and demonstrated by benchmarking the state-of-the-art TinyML frameworks TFLite for Microcontrollers and TVM effortlessly with a large number of configurations in a low amount of time.
\end{abstract}


\section{Introduction}
TinyML is one of the current challenges in the embedded software and hardware technology and hardware business, confronting tiny edge devices with machine learning tasks. These chips, which are sometimes even connected to the cloud, are not only limited in their processing capabilities and memory capacity, but also operate on a very small power budget due to size and cost constraints.

\subsection{Motivation}

Optimizing TinyML applications to perform better or run more efficiently requires considerations at several stages during the design process. Of course, this starts with the model design itself. However, the  deployment method, as well as the used hardware design, should not be underestimated from the beginning. Benchmarking solutions help to decide which approaches should be used for a given application and the possibility of virtual prototyping at early design stages can provide relevant estimates of the final performance even before actual hardware is available. Unfortunately, existing benchmarking tools are often limited to a specific application, framework or set of target devices, which makes comparing the available TinyML tools and methods more difficult. The MLonMCU project proposed in this paper solves this issue by providing a framework-independent, easily extensible and powerful benchmarking solution which also offers fast retargeting possibilities.
We were able to generate 118 end-to-end comparisons with minimal effort in under 60 minutes.

\subsection{State-of-the-Art}

The term TinyML
was first referenced in \cite{warden2020tinyml} and is nowadays well-accepted in the industry and research community. The number of TinyML frameworks is growing, with \emph{TensorFlow Lite for Microcontrollers (TFLM)}   currently being the  most relevant \cite{david2021tensorflow}. It provides a more lightweight version of the well-established TFLite framework, which is frequently used on mobile devices \cite{louis2019towards}. For extreme edge applications, where very limited amount of memory is available, \emph{TFLite Micro Compiler}, a code generation tool reducing the memory overheads by generating fully static inference code, was proposed in \cite{stahl2021tflmc}.
The open source deep learning optimization framework TVM was proposed in \cite{chen2018tvm} and allows leveraging compiler-like optimization methods for machine learning models.
MicroTVM, an addition to the TVM framework targeting bare metal devices, was introduced to deal with common deployment challenges on these devices. The standardized TinyML benchmarks used in this work are proposed in \cite{banbury2021mlperf}.

\section{Implementation}

The tool implemented in the context of this paper
has the main goal to enable performing extensive benchmarks on TinyML models, frameworks and targets.
Further, the core design principles are:

\begin{itemize}
    \item Isolation: The utilities used by MLonMCU should not interfere with any other programs running on a system.
    \item Reproducibility: All intermediate artifacts of a benchmarking session should be made available to the user.
    \item Parallelism:
    MLonMCU should use all the available computational resources to deliver results as fast as possible.
    \item Extensibility: Custom user-written code should integrate easily with the existing MLonMCU codebase.
\end{itemize}

\begin{figure}[h!]
    \centering
    \includegraphics[width=.99\columnwidth]{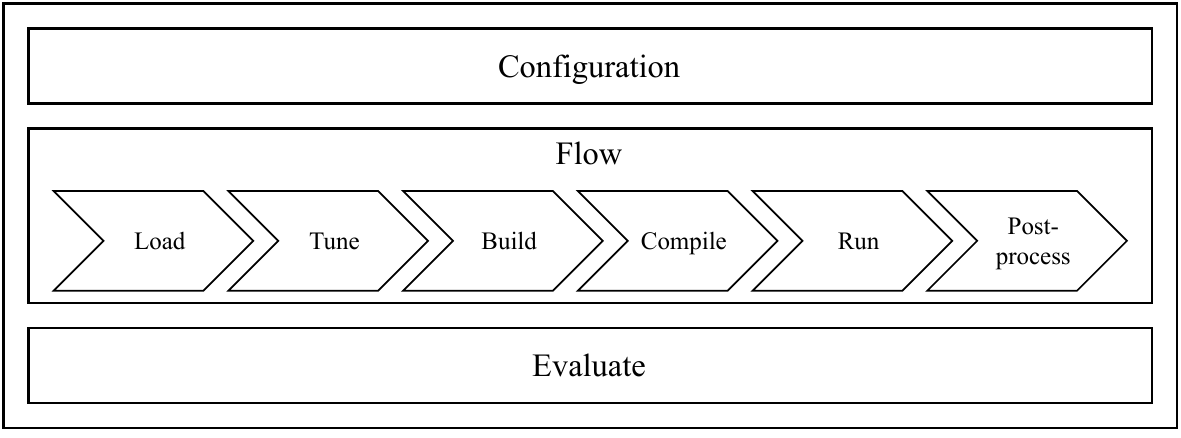}
    \caption{Structure of MLonMCU.}
    \label{fig:mlonmcu_structure}
\end{figure}

\subsection{Overview}

The project which is available on GitHub\footnote{Open-source repository: \url{https://github.com/tum-ei-eda/mlonmcu}} was implemented in Python and consists of three main modules whose roles are depicted in Fig.~\ref{fig:mlonmcu_structure}. A straightforward to use but very powerful command line interface can be utilized in addition to the provided Python development interface to interact with each of MLonMCU's components.

\subsubsection{Configuration}
A prerequisite step to using MLonMCU's core features is initializing and installing at least one environment. Predefined environment templates are supplied.
The idea of allowing to install several environments in parallel is motivated by the need for isolated dependencies and reproducibility.

\subsubsection{Flow}

The heart of MLonMCU is the definition and execution of single benchmarks or complex benchmarking sessions. Each benchmark consists of a set of stages (see Fig.~\ref{fig:mlonmcu_structure}) which will be invoked until the final stage is reached.

\subsubsection{Evaluate}

Each session generates a set of artifacts, which can be used to  further investigate the code generation results or to debug target-related problems, as well as a report with several details about each run. These metrics may consist of the model execution latency (e.g. Instructions/Cycles/Runtime) as well as static (and optionally dynamic) memory usage information.
Using the included Python development API, these reports and artifacts can directly be used to do further analysis of the data or to create comprehensive visualizations.

\subsection{Components}\label{sec:components}

In the following, the types of components supported by the MLonMCU tool and their relation to the mentioned stages in Fig.~\ref{fig:mlonmcu_structure} are briefly introduced.

\subsubsection{Frontends}

Given a model name or path, during the \emph{Load} stage, an automatically chosen frontend ensures that the model is made available to all the upcoming stages.

\subsubsection{Frameworks and Backends}

For each framework supported by MLonMCU, one or more backends are available. The role of those backends during the \emph{Build} stage is converting a provided model file into inference code, including a consistent interface for the target software cross-compiled in the following stage.
If supported by the chosen framework, a \emph{Tune} stage can be added to the flow as well.

\subsubsection{Platforms and Targets}

MLonMCU follows different concepts to handle various kinds of target devices or simulators: The supported simulators are directly managed by MLonMCU using target-specific routines for compiling and running generated programs. While this introduces a great degree of control, it does not scale well to support a large number of devices, especially if actual hardware is involved. For this reason, the complex handling of different toolchains and flashing via serial ports can be taken care by platforms (e.g. the Zephyr project)
designated for such purpose. This  allows MLonMCU to target a large number of devices ``out of the box''. A target software library called \textit{Machine Learning Interface} is used as an abstraction layer standardizing how models are executed with MLonMCU and benchmark results are reported by connected devices in a platform-independent fashion. The platforms are involved in the \emph{Compile} and \emph{Run} stage of the flow.

\subsubsection{Postprocesses}

Predefined procedures can be invoked in the final stage. Their usage is not  limited to applying transformations like filters on the resulting report, as also any previously generated artifacts can be considered as well. A combination of some postprocesses can reduce the size of the final report to contain only the relevant information and automatically generate visualization artifacts such as graphs.

\subsubsection{Features}

A special type of component are features as they affect the way how other components interact with each other. For each type of component there exist a feature base class providing utilities to overwrite individual components' configurations. One of these features allows comparing the inference outputs against previously defined ``golden'' reference values which is useful to detect if a framework degrades the models' accuracy.

\section{Evaluation}

To demonstrate the usefulness of the implemented tool, a set of TinyML deployment-related benchmarking questions are answered in this section. The raw data used to create each discussed table can be obtained using a single invocation of the MLonMCU command line interface.
The two main points of discussion are the overheads introduced by several supported framework backends in MLonMCU as well as the runtime performance of various TVM schedules, which define transformations of the computations in a program, on resource-constrained microcontroller hardware. 

\subsection{Methodology} 
In the following, the methodology is explained by introducing the underlying benchmarks and used devices. 

\subsubsection{Models}

The MLPerf Tiny benchmark was utilized \cite{banbury2021mlperf} for all evaluations in Sections \ref{sec:experiment1} and \ref{sec:experiment2}. Using the $\SI{8}{\bit}$ quantized integer variants instead of the $\SI{32}{\bit}$ floating-point models is an evident choice when dealing with resource-constrained hardware, especially as the quantization often introduces no loss in accuracy. The four models are listed in Table~\ref{table:mlperftiny} with their individual use cases and file sizes. Three of them are heavily relying on a number of (depthwise-separable) convolutional and pooling layers in a CNN architecture, while the anomaly detection model \verb|toycar| uses a traditional DNN.

\begin{table}[h!]
\centering
\caption{MLPerf Tiny Benchmark Models.}
\label{table:mlperftiny}
\begin{tabular}{||c c c||} 
 \hline
 Name & Use Case & Quantized Size \\ [0.5ex] 
 \hline\hline
 \verb|aww| & Keyword Spotting & $\SI{58.3}{\kilo\byte}$ \\ 
 \verb|vww| & Visual Wake Words & $\SI{325}{\kilo\byte}$ \\
 \verb|resnet| & Image Classification & $\SI{96.2}{\kilo\byte}$ \\
 \verb|toycar| & Anomaly Detection & $\SI{270}{\kilo\byte}$ \\
 \hline
\end{tabular}

\end{table}

\subsubsection{Targets}\label{sec:targets}

For the backend comparisons in Section \ref{sec:experiment1}, it is convenient to use an instruction set simulator (ISS). Thus, the ETISS target supported by MLonMCU is utilized to simulate a $\SI{32}{\bit}$ RISC-V microcontroller with an RV32GC
(RV32IMAFDC)
core \cite{mueller2017extendable}.
Afterwards, in Section \ref{sec:experiment2}, the previously introduced models are deployed on microcontroller hardware to compare how the TVM framework can adapt to the inherent differences in their microarchitecture.

\begin{table}[h!]
\centering
\caption{Used hardware targets.}
\label{table:hwtargets}
\begin{tabular}{||c c c c c||} 
 \hline
 Name & Architecture & Clock & Flash & (S)RAM \\ [0.5ex] 
 \hline\hline
 \verb|esp32c3| & RV32IMC & $\SI{160}{\mega\Hz}$ & $\SI{2}{\mega\byte}$ & $\SI{384}{\kilo\byte}$ \\[0.5em] 
 \verb|stm32f4| & ARM Cortex-M4 & $\SI{100}{\mega\Hz}$ & $\SI{1.5}{\mega\byte}$ &  $\SI{320}{\kilo\byte}$ \\[0.5em] 
 \verb|stm32f7| & ARM Cortex-M7 & $\underset{\text{(dual issue)}}{\SI{216}{\mega\Hz}}$ & $\SI{2}{\mega\byte}$ & $\SI{512}{\kilo\byte}$\\[0.5em] 
  \verb|esp32| & Xtensa LX6 & $\SI{240}{\mega\Hz}$ & $\SI{448}{\kilo\byte}$ & $\SI{328}{\kilo\byte}$ \\
 \hline
\end{tabular}

\end{table}

The list of used targets given in Table~\ref{table:hwtargets} covers three different instruction set architectures (ISAs) of two different chip vendors. The available flash memory constrains the size of the models deployed on these targets. However, most of the time, the available (S)RAM for storing intermediate activations of the network is a more limiting factor. The microcontrollers are using only a single core clocked between $\SI{100}{}$ and $\SI{240}{\mega\Hz}$ for model execution. Surveying the usefulness of the utilized  hardware should not be the main point of discussion. Thus, target-side optimizations such as ISA extensions or specialized kernel libraries are not considered here.

\subsubsection{Environment}\label{sec:environment}

The following results have been generated using MLonMCU on a quad-core x86 CPU\footnote{Intel(R) Core(TM) i7-6700 CPU @ 3.40GHz} in about $50$ minutes. In Table~\ref{table:runtime} the total runtime (excluding tuning time) is given for both benchmarks.

\begin{table}[h!]
\centering
\caption{Benchmark runtime summary.}
\label{table:runtime}
\begin{tabular}{||c c c c||} 
 \hline
 \multirow{2}{*}{Benchmark} &  \multirow{2}{*}{\#Runs} & \multicolumn{2}{c||}{Runtime}\\[0.25em] 
            &       & Load - Compile & Load - Run \\
 \hline\hline
 \ref{sec:experiment1} & 20 & $\SI{340}{\sec}$ & $\SI{350}{\sec}$\\
 \ref{sec:experiment2} & 98 & $\approx\SI{16}{\minute}$ & $\approx\SI{43}{\minute}$\\

 \hline
\end{tabular}
\end{table}

It is evident that the total runtime for the experiments on real hardware (Benchmark C) is dominated by factors which can not be optimized directly by MLonMCU, e.g. the time required for flashing the target software binary and running the actual program. Another interesting observation is the difference in mean build times ($\SI{17}{\frac{\sec}{\text{Run}}}$ vs. $\SI{9}{\frac{\sec}{\text{Run}}}$) between the two sets of benchmarks which can be explained by the fact that compiling the target software for TVM is much faster compared to TFLM (only used in the first experiment) due to the lower number of source files involved in the compilation.

\begin{table*}[t]
\centering
\caption{Backend comparisons.}
\label{table:backends}
\begin{tabular}{||c|c|c|c||c|c|c|c||} 
 \hline
 \multirow{2}{*}{Model} & \multirow{2}{*}{Metric} & \multicolumn{2}{c||}{TFLM} & \multicolumn{3}{c|}{TVM} & \multirow{2}{*}{Unit}\\\cline{3-7}
                        & &  \verb|tflmi| & \verb|tflmc| & \verb|tvmaot| & \verb|tvmaot+| & \verb|tvmrt| &\\
 \hline\hline
 \multirow{4}{*}{\texttt{aww}}  & \#Instr. (Setup) & 264 & 62 ($-76\%$) & $\approx\mathbf{0}$ &  $\approx\mathbf{0}$ & 2988 & $\times 10^3$\\
                        & \#Instr. (Invoke) & 153.144 & 153.143 ($\pm 0\%$) & \textbf{29.819} & 30.671 ($+2.5\%$) & 33.660 ($+2.9\%$)& $\times 10^6$\\
         & ROM  & 143 & \textbf{107} ($-24.8\%$) & 126 & 122 ($-3\%$) & 164 ($+30\%$)& $\SI{}{\kilo\byte}$\\
         & RAM & 37 & \textbf{28} ($-24.5\%$) & 174 & 125 ($-28.3\%$) & 1056 ($+605\%$)& $\SI{}{\kilo\byte}$\\\hline
 \multirow{4}{*}{\texttt{vww}}& \#Instr. (Setup)  & 1025 & 274 ($-73\%$) & $\approx\mathbf{0}$ & $\approx\mathbf{0}$ & 10688 & $\times 10^3$\\ 
         & \#Instr. (Invoke) & 432.031 & 432.028 ($\pm 0\%$) & 89.672 & \textbf{87.460} ($-2.5\%$) & 91.885 ($+2.5\%$) & $\times 10^6$\\
         & ROM  & 416 & \textbf{342} ($-17.8\%$) & 579 & 571 ($-1.4\%$) & 655 ($+113\%$) & $\SI{}{\kilo\byte}$\\
         & RAM & 337 & \textbf{274} ($-17.8\%$) & 496 & 495 ($-0.2\%$) & 4229 ($+853\%$) & $\SI{}{\kilo\byte}$\\\hline
 \multirow{4}{*}{\texttt{resnet}}& \#Instr. (Setup)  & 217 & 41 ($-81\%$) & $\approx\mathbf{0}$ & $\approx\mathbf{0}$ & 3970 & $\times 10^3$\\
         & \#Instr. (Invoke) & 687.462 & 687.45 ($\pm 0\%$) & \textbf{114.802} & 116.115 ($+1.1\%$) & 115.671 ($+0.8\%$) & $\times 10^6$\\
         & ROM  & 183 & \textbf{160} ($-12.6\%$) & 228 & 224 ($-1.8\%$) & 274 ($+20.2\%$) & $\SI{}{\kilo\byte}$\\
         & RAM & 69 & \textbf{58} ($-15.9\%$) & 125 & 108 ($-13.6\%$) & 1055 ($+844\%$) & $\SI{}{\kilo\byte}$\\\hline
 \multirow{4}{*}{\texttt{toycar}}& \#Instr. (Setup)  & 71 & 5 ($-92\%$) & $\approx\mathbf{0}$ & $\approx\mathbf{0}$ & 5014 & $\times 10^3$\\
         & \#Instr. (Invoke) & 3.001 & 2.996 ($-1.6\%$) & \textbf{2.441} & 2.457 ($+0.6\%$) & 2.442 ($\pm 0\%$) & $\times 10^6$\\
         & ROM  & 345 & \textbf{330} ($-4.3\%$) & 594 & 592 ($-0.3\%$) & 631 ($+10.6\%$) & $\SI{}{\kilo\byte}$\\
         & RAM & 21 & \textbf{7} ($-63\%$) & 8 & \textbf{7} ($-8.9\%$) & 1057 ($+14,374\%$) & $\SI{}{\kilo\byte}$\\\hline
\end{tabular}

\end{table*}

\subsection{Comparison of TinyML backends and their runtime overhead}\label{sec:experiment1}

First, the available backends in MLonMCU are compared with each other using the data given in Table~\ref{table:backends}. After an evaluation of the individual backends of each frameworks, the overall performance of the underlying frameworks TFLM and TVM is considered as well.

The default way to deploy a machine learning model using \emph{TensorFlow Lite for Microcontrollers} is based on the \emph{TFLite Micro Interpreter} (\texttt{tflmi}), which parses a TFLite ``FlatBuffer'' data structure at runtime \cite{david2021tensorflow}. An alternative approach is available using the \emph{TFLite Micro Compiler} (\texttt{tflmc}) project proposed in \cite{stahl2021tflmc}, which generates minimal inference code for a given model. As expected, a reduction of ROM usage between $15$ and $\SI{30}{\kilo\byte}$ can be achieved by eliminating the code-size overhead to implement the interpreter.
A reduction in RAM usage of at least $12\%$ can be expected, too.
Because both backends loop over the same set of kernels, their inference performance is equivalent, while the one-time initialization time of the models can be reduced by utilizing the \texttt{tflmc} backend. This setup time is typically orders of magnitudes smaller than the required time to invoke a model, thus it can be neglected as soon as multiple inferences should be run.

TVM supports two main approaches to deploy the generated kernels for a given model on an edge device. The \emph{Graph Executor} (\texttt{tvmrt}) is following a similar approach as the \texttt{tflmi} backend, parsing a JSON representation of the model at runtime. This JSON parser and other components introduce a code size overhead of about $\SI{40}{\kilo\byte}$.
An alternative, more minimalistic approach is called \emph{Ahead-of-Time  Executor} (\texttt{tvmaot}). It also generates the top-level inference function, outperforming \texttt{tvmrt} in every considered metric, but especially in terms of RAM overhead.
A third backend named \texttt{tvmaot+} 
is provided by MLonMCU, enabling the recently implemented Unified Static Memory Planner (USMP) and further runtime-related optimizations in addition to the default set of features. This can reduce the RAM usage for three of the four models by $9$ to $28\%$. The AoT-compiled models basically have no initialization steps, while the \texttt{tvmrt} requires at least one million instructions to prepare for the model execution, exceeding even the inference time for less complex models such as \texttt{toycar}. While the \texttt{tvmrt} backend introduces several overheads regarding the discussed performance and memory metrics, it is still a very powerful tool, as it allows profiling the model execution on the target device and provides the necessary utilities to use AutoTVM with MicroTVM workloads.

The metrics of the best-performing backend in Table~\ref{table:backends} are given in bold digits. In terms of inference performance, the default kernel implementations provided by TFLite Micro can not keep up with TVM's auto-generated kernels. This makes TVM an obvious choice if the optimal inference latency should be reached.
In terms of ROM and RAM usage, TFLM outperforms TVM for more complex models, often even by a factor of two. This behavior can be explained by a legalization pass upcasting all $\SI{8}{\bit}$ tensors to $\SI{16}{\bit}$ data types, which is not desirable when dealing with a memory-constrained target.
If inference performance using TFLM is sufficient for a given application, the \texttt{tflmc} backend can also be considered to deploy a model with minimal memory overheads.

\subsection{Evaluation of TVM schedules on microcontroller hardware}\label{sec:experiment2}

In the following, the four MLPerf Tiny models are deployed on the four different targets introduced in Section \ref{sec:targets} using MLonMCU's Zephyr platform. For each of these 16 combinations, up to eight different types of TVM schedules are compared, resulting in about 100 benchmarks results given in Table~\ref{table:tvm_schedules}. The measured inference time is given in seconds while failing benchmarks due to insufficient available memory are indicated by a $-$.
The most complex model in terms of execution time is \texttt{resnet}, followed by \texttt{vww}, \texttt{aww} and finally \texttt{toycar}, matching the order of instruction counts given in Table~\ref{table:backends}. While the targets \texttt{esp32c3} and \texttt{stm32f7} have been able to run all four models without reaching memory limits, both \texttt{stm32f4} and \texttt{esp32} failed to deploy the large visual wake-word network, at least for some schedules, due to insufficient amount of RAM available.

Only considering untuned results, it can be stated: The choice of the used data layout\footnote{NHWC: Channels-last (TFLite default), NCHW: Channels-first (TVM default)} has a large impact on the measured inference performance, especially for the \texttt{vww} and \texttt{resnet} model on the \texttt{esp32c3} and \texttt{esp32} target, while for the rest, the difference in inference latency is between $\times 1.5$ and $\times 2$ making the channels-first layout (NCHW) a better choice for these embedded targets. When using the NCHW layout in TVM, the activation and kernel tensors are internally transformed into a 5- respectively 6-dimensional NCHWc (OIHWio) layout to improve spacial locality leading to the large gap in inference time between the two considered layouts.

In addition to TVM's default schedules (mainly targeting x86 architectures), operator implementations intended for usage with larger ARM (Aarch64) targets are now considered as well. It can be seen that on most CNNs those kernels perform similar or worse than TVM's default implementations for both types of layouts while for the only DNN the dense/fully-connected operators for ARM targets are able to run up two times more efficiently. 
If leveraging the AutoTVM feature, benchmarking several parameterized operator implementations on the actual target device to find the one which performs best, further observations can be made. For each of the supported targets, a second column is available in Table~\ref{table:tvm_schedules} providing the model execution time after tuning the network. The tuning was performed beforehand for at least 600 iterations per combination.
The impact of auto-tuning depends heavily on the used schedules and layouts.
For x86 NHWC schedules, only fully-connected layers are tunable, leading to negligible results on CNNs, which are typically dominated by the convolutional layers. Tunable convolution schedules exist for both considered NCHW schedules as well as for NHWC schedules written for ARM targets, allowing to optimize CNNs effectively to improve the inference performance. Finally, it turns out that no tuning-templates for fully-connected operator implementations on ARM targets have been written so far, leading to zero improvements in the last row of Table~\ref{table:tvm_schedules}.

For each combination of targets and models, the best-performing result is highlighted in Table~\ref{table:tvm_schedules}. For CNNs, TVM's default NCHW schedules performed best, especially with autotuning-enabled, while for DNNs such as the \texttt{toycar} network, ARM schedules are a better choice.
It is likely that even more improvements can be achieved by increasing the number of tuning iterations. However, this becomes a non-trivial task for TinyML devices as MicroTVM currently needs to cross-compile, flash and run a new program for every single tuning iteration, which is very time intensive and also degrades the lifetime of the flash memory used by the microcontrollers.

The massive improvements in inference latency between different target architectures can often be explained by the used ARM compiler which seems to be more sophisticated compared to the other ones.
While the \texttt{esp32c3} and \texttt{esp32} share the same board vendor, they are based on two different instruction set architectures. The \texttt{esp32} is clocked $50\%$ higher than the newer \texttt{esp32c3} leading to similar or better performance in most of the rows.

\begin{table*}[t]
\centering
\caption{TVM schedules on different target hardware.}
\label{table:tvm_schedules}
\begin{tabular}{||c|c|cc||cc|cc||cc||} 
 \hline
 \multirow{2}{*}{Model} & \multirow{2}{*}{Schedules (Layout)} & \multicolumn{2}{c||}{RISC-V} & \multicolumn{4}{c||}{ARM (Cortex-M)} & \multicolumn{2}{c||}{Xtensa (LX6)} \\\cline{3-10}
                        & & \multicolumn{2}{c||}{\texttt{esp32c3}} & \multicolumn{2}{c|}{\texttt{stm32f4}} & \multicolumn{2}{c||}{\texttt{stm32f7}} & \multicolumn{2}{c||}{\texttt{esp32}} \\
 \hline\hline
     & AutoTVM? & no & yes & no & yes & no & yes & no & yes  \\\hline
 \multirow{4}{*}{\texttt{aww}}  & Default (NHWC) & $\SI{0.210}{\sec}$ & $\SI{0.209}{\sec}$ & $\SI{0.302}{\sec}$ & $\SI{0.302}{\sec}$ & $\SI{0.065}{\sec}$ & $\SI{0.065}{\sec}$ & $\SI{0.136}{\sec}$ &
 $-$ \\
                      & Default (NCHW) & $\SI{0.113}{\sec}$ & \textbf{$\SI{0.092}{\sec}$} & $\SI{0.220}{\sec}$ & $-$ & $\SI{0.043}{\sec}$ & \textbf{$\SI{0.029}{\sec}$} & \textbf{$\SI{0.125}{\sec}$} & $-$\\
           & ARM (NHWC)  & $\SI{0.248}{\sec}$  & $\SI{0.284}{\sec}$ & $\SI{0.203}{\sec}$ & $-$ & $\SI{0.084}{\sec}$ & $\SI{0.052}{\sec}$ & $\SI{0.159}{\sec}$ & $-$\\
         & ARM (NCHW)  & $\SI{0.161}{\sec}$ & $\SI{0.144}{\sec}$ & $\SI{0.29}{\sec}$ & \textbf{$\SI{0.163}{\sec}$} & $\SI{0.067}{\sec}$ & $\SI{0.063}{\sec}$ & $\SI{0.155}{\sec}$ & $-$\\\hline
 \multirow{4}{*}{\texttt{vww}} & Default (NHWC) & $\SI{16.037}{\sec}$ & $\SI{16.035}{\sec}$ & $-$ &  $-$ & $\SI{0.336}{\sec}$ & $\SI{0.336}{\sec}$ & $-$ &
 $-$\\
                     & Default (NCHW)  & $\SI{0.349}{\sec}$ & \textbf{$\SI{0.292}{\sec}$} & \textbf{$\SI{0.395}{\sec}$} & $-$ & $\SI{0.127}{\sec}$ & \textbf{$\SI{0.094}{\sec}$} & $-$ & $-$\\
         & ARM (NHWC)  & $\SI{17.019}{\sec}$ & $\SI{16.03}{\sec}$ & $\SI{0.555}{\sec}$ & $\SI{0.474}{\sec}$ & $\SI{0.429}{\sec}$ & $\SI{0.173}{\sec}$ & $-$ & $-$\\
          & ARM (NCHW) & $\SI{0.482}{\sec}$ & $\SI{0.430}{\sec}$ & $\SI{0.855}{\sec}$ & $\SI{0.469}{\sec}$ & $\SI{0.209}{\sec}$ & $\SI{0.188}{\sec}$ & $-$ & $-$\\\hline
 \multirow{4}{*}{\texttt{resnet}}  & Default (NHWC)& $\SI{24.729}{\sec}$ & $\SI{24.728}{\sec}$ & $\SI{0.974}{\sec}$ & $\SI{0.974}{\sec}$ & $\SI{0.455}{\sec}$ & $\SI{0.455}{\sec}$  & $\SI{11.707}{\sec}$ &
 $-$\\
                       & Default (NCHW) & $\SI{0.397}{\sec}$ & \textbf{$\SI{0.300}{\sec}$} & $\SI{0.424}{\sec}$ & \textbf{$\SI{0.385}{\sec}$} & $\SI{0.158}{\sec}$ & \textbf{$\SI{0.108}{\sec}$} & \textbf{$\SI{0.446}{\sec}$} & $-$\\
          & ARM (NHWC)  & $\SI{25.541}{\sec}$ & $\SI{2.146}{\sec}$  & $\SI{1.237}{\sec}$ & $\SI{0.522}{\sec}$ & $\SI{0.564}{\sec}$ & $\SI{0.191}{\sec}$ & $\SI{12.22}{\sec}$ & $-$\\
         & ARM (NCHW)  & $\SI{0.551}{\sec}$ & $\SI{0.550}{\sec}$ & $\SI{0.968}{\sec}$ & $\SI{0.612}{\sec}$ & $\SI{0.295}{\sec}$ & $\SI{0.257}{\sec}$ & $\SI{0.733}{\sec}$ & $-$\\\hline
 \multirow{2}{*}{\texttt{toycar}}  & Default  & $\SI{0.075}{\sec}$ & $\SI{0.073}{\sec}$ & $\SI{0.029}{\sec}$ &  $\SI{0.023}{\sec}$ & $\SI{0.012}{\sec}$ & \textbf{$\SI{0.003}{\sec}$} & $\SI{0.078}{\sec}$ & $-$\\
          & ARM  & \textbf{$\SI{0.04}{\sec}$} & \textbf{$\SI{0.04}{\sec}$} & \textbf{$\SI{0.019}{\sec}$} & \textbf{$\SI{0.019}{\sec}$} & $\SI{0.007}{\sec}$ & $\SI{0.007}{\sec}$ & \textbf{$\SI{0.047}{\sec}$} & $-$ \\\hline
\end{tabular} 

\end{table*}

\section{Conclusion}
The designed MLonMCU tool solves challenges with benchmarking of TinyML applications by automating several steps in the deployment flow in a straightforward fashion. Tasks such as the comparison of TinyML frameworks or different hardware targets can be accomplished effortlessly as demonstrated in the previous sections. 
Especially using the TVM ML compiler suite, promising results in terms of inference performance and deployment overheads on edge devices have been observed.
Limitations in terms of usability of both discussed frameworks have been discussed as well.

Future work can build up on MLonMCU's infrastructure, e.g. to incorporate model deployment specific metrics in Network Architecture Search (NAS) algorithms
to find optimal models for a given target. The impact of custom kernel libraries such as CMSIS-NN \cite{lai2018cmsis}, ISA extensions and hardware accelerators is of high interest and should be investigated with MLonMCU, as well. A study of the power consumption of the previously discussed workloads on a broader field of devices may supplement the generated results in the future.

\bibliographystyle{IEEEtran}
\bibliography{IEEEabrv,IEEEReferences}

\begin{thebibliography}{1}
\providecommand{\url}[1]{#1}
\csname url@samestyle\endcsname
\providecommand{\newblock}{\relax}
\providecommand{\bibinfo}[2]{#2}
\providecommand{\BIBentrySTDinterwordspacing}{\spaceskip=0pt\relax}
\providecommand{\BIBentryALTinterwordstretchfactor}{4}
\providecommand{\BIBentryALTinterwordspacing}{\spaceskip=\fontdimen2\font plus
\BIBentryALTinterwordstretchfactor\fontdimen3\font minus
  \fontdimen4\font\relax}
\providecommand{\BIBforeignlanguage}[2]{{%
\expandafter\ifx\csname l@#1\endcsname\relax
\typeout{** WARNING: IEEEtran.bst: No hyphenation pattern has been}%
\typeout{** loaded for the language `#1'. Using the pattern for}%
\typeout{** the default language instead.}%
\else
\language=\csname l@#1\endcsname
\fi
#2}}
\providecommand{\BIBdecl}{\relax}
\BIBdecl

\bibitem{warden2020tinyml}
\BIBentryALTinterwordspacing
P.~Warden and D.~Situnayake, \emph{TinyML: Machine Learning with TensorFlow
  Lite on Arduino and Ultra-low-power Microcontrollers}.\hskip 1em plus 0.5em
  minus 0.4em\relax O'Reilly, 2019. [Online]. Available:
  \url{https://books.google.de/books?id=sB3mxQEACAAJ}
\BIBentrySTDinterwordspacing

\bibitem{david2021tensorflow}
R.~David, J.~Duke, A.~Jain, V.~Janapa~Reddi, N.~Jeffries, J.~Li, N.~Kreeger,
  I.~Nappier, M.~Natraj, T.~Wang \emph{et~al.}, ``Tensorflow lite micro:
  Embedded machine learning for tinyml systems,'' \emph{Proceedings of Machine
  Learning and Systems}, vol.~3, pp. 800--811, 2021.

\bibitem{louis2019towards}
M.~S. Louis, Z.~Azad, L.~Delshadtehrani, S.~Gupta, P.~Warden, V.~J. Reddi, and
  A.~Joshi, ``Towards deep learning using tensorflow lite on risc-v,'' in
  \emph{Third Workshop on Computer Architecture Research with RISC-V (CARRV)},
  vol.~1, 2019, p.~6.

\bibitem{stahl2021tflmc}
\BIBentryALTinterwordspacing
R.~Stahl, ````exploring static code generation and simd-acceleration for
  machine learning on risc-v'' in in ``risc-v forum: Developer tools \& tool
  chains'','' 2021, [Recording available: \url{https://youtu.be/NLGAjdVIzkk}].
  [Online]. Available: \url{https://riscvforumdttc2021.sched.com/event/jGkT}
\BIBentrySTDinterwordspacing

\bibitem{chen2018tvm}
T.~Chen, T.~Moreau, Z.~Jiang, L.~Zheng, E.~Yan, H.~Shen, M.~Cowan, L.~Wang,
  Y.~Hu, L.~Ceze \emph{et~al.}, ``$\{$TVM$\}$: An automated $\{$End-to-End$\}$
  optimizing compiler for deep learning,'' in \emph{13th USENIX Symposium on
  Operating Systems Design and Implementation (OSDI 18)}, 2018, pp. 578--594.

\bibitem{banbury2021mlperf}
C.~Banbury, V.~J. Reddi, P.~Torelli, J.~Holleman, N.~Jeffries, C.~Kiraly,
  P.~Montino, D.~Kanter, S.~Ahmed, D.~Pau \emph{et~al.}, ``Mlperf tiny
  benchmark,'' \emph{arXiv preprint arXiv:2106.07597}, 2021.

\bibitem{mueller2017extendable}
D.~Mueller-Gritschneder, M.~Dittrich, M.~Greim, K.~Devarajegowda, W.~Ecker, and
  U.~Schlichtmann, ``The extendable translating instruction set simulator
  (etiss) interlinked with an mda framework for fast risc prototyping,'' in
  \emph{2017 International Symposium on Rapid System Prototyping (RSP)}.\hskip
  1em plus 0.5em minus 0.4em\relax IEEE, 2017, pp. 79--84.

\bibitem{lai2018cmsis}
L.~Lai, N.~Suda, and V.~Chandra, ``Cmsis-nn: Efficient neural network kernels
  for arm cortex-m cpus,'' \emph{arXiv preprint arXiv:1801.06601}, 2018.

\end{thebibliography}

\end{document}